\documentclass{article}
\usepackage{spconf,amsmath,graphicx}
\usepackage{amssymb}
\usepackage{booktabs}

\title{ADVERSARIAL JAMMING FOR AUTOENCODER DISTRIBUTION MATCHING}
\name{Waleed El-Geresy and Deniz Gündüz\thanks{This work received funding from the UKRI for the projects AI-R (ERC-Consolidator Grant, EP/X030806/1) and CONNECT (EP/T023600/1).}}
\address{Imperial College London}
\begin{document}
\maketitle
\begin{abstract}
We propose the use of adversarial wireless jamming to regularise the latent space of an autoencoder to match a diagonal Gaussian distribution. We consider the minimisation of a mean squared error distortion, where a jammer attempts to disrupt the recovery of a Gaussian source encoded and transmitted over the adversarial channel. A straightforward consequence of existing theoretical results is the fact that the saddle point of a minimax game - involving such an encoder, its corresponding decoder, and an adversarial jammer - consists of diagonal Gaussian noise output by the jammer. We use this result as inspiration for a novel approach to distribution matching in the latent space, utilising jamming as an auxiliary objective to encourage the aggregated latent posterior to match a diagonal Gaussian distribution. Using this new technique, we achieve distribution matching comparable to standard variational autoencoders and to Wasserstein autoencoders. This approach can also be generalised to other latent distributions.
\end{abstract}
\section{Introduction}
\label{sec:intro}

\begin{figure}[t]
    \centering
    \includegraphics[width=\linewidth]{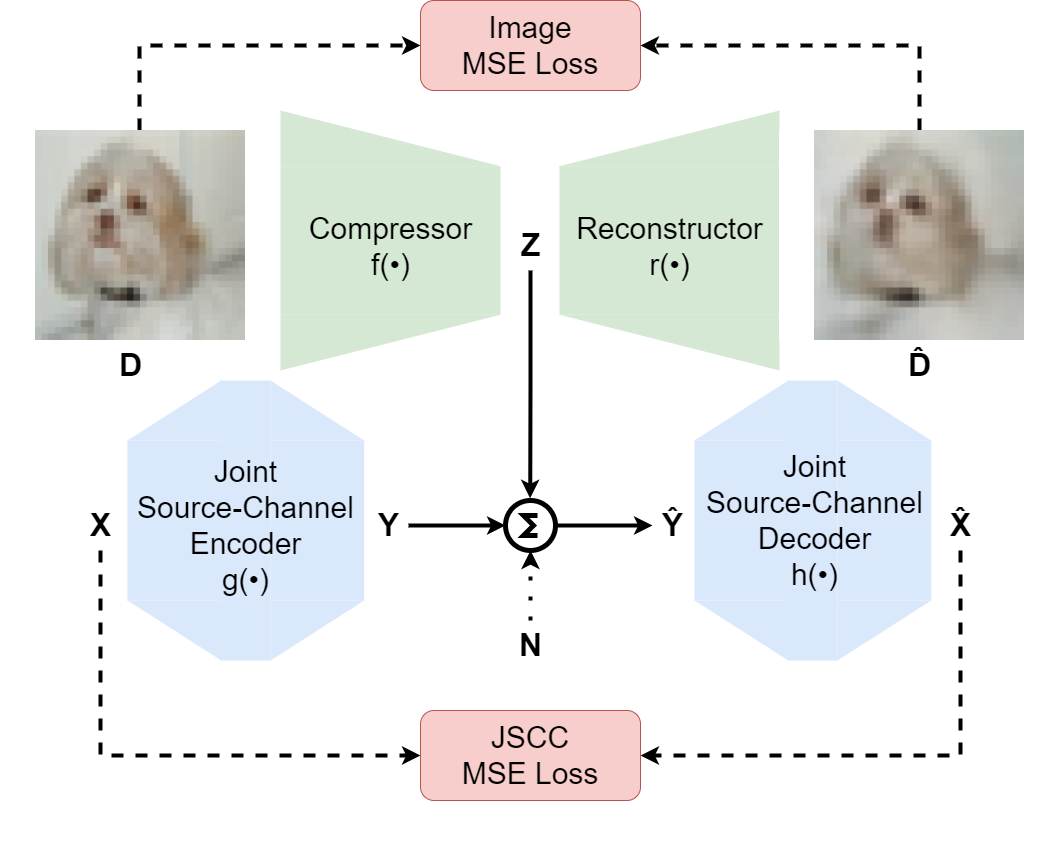}
    \caption{Adversarial jamming for prior distribution matching. The game between a compressor/jammer \(f\) (compressing the images) and a JSCC autoencoder \((g, h)\) (for source \(\mathbf{X} \sim \mathcal{N}(\mathbf{0}, \mathbf{I})\)) has a saddle point with \(Z^* \sim \mathcal{N}(\mathbf{0}, \mathbf{I})\), implicitly imposing this prior on the latent space.}
    \label{fig:jscc_scheme}
\end{figure}

Generative modelling can be described as the task of drawing artificial samples from an underlying distribution based on a the observation of a real set of samples from that distribution. The distribution can be simple and tractable, or it can be complex and intractable. A variety of methods for generative modelling in the field of machine learning and deep learning have been proposed. These include diffusion models \cite{hoDenoisingDiffusionProbabilistic2020}, generative adversarial networks (GANs) \cite{goodfellowGenerativeAdversarialNetworks2014}, variational autoencoders (VAEs) \cite{kingmaAutoEncodingVariationalBayes2014}, normalising flows \cite{dinhDensityEstimationUsing2016}, and energy-based models \cite{bond-taylorDeepGenerativeModelling2022}. Many of these can generate samples from the modelled distribution given a representation (invert the mapping). However, the ability to map a sample (either real or generated) back to an appropriate representation is unique to symmetric techniques, namely VAEs and normalising flows.

Here, we focus on VAEs, \cite{kingmaAutoEncodingVariationalBayes2014} which use a variational approximation to model an intractable prior distribution for a particular data distribution. Commonly, a parameterised multivariate Gaussian distribution is used as the prior. This is because the Kullback-Leibler (KL) divergence - in the case of a multivariate Gaussian - has an analytic differentiable form, which can be used as part of the loss function. Secondly, using a known distribution such as a multivariate Gaussian can allow us to sample from the data distribution by first sampling this prior and then generating samples from the resultant posterior using the autoencoder's decoder.

Existing work in the field of VAEs involving the matching of the aggregated posterior to a known prior, such as a diagonal Gaussian distribution, has consisted of a number of methods. The initial VAE relied on maximising the so-called evidence lower bound (ELBO) - a lower bound on the likelihood of generated data coming from the true distribution, used as a proxy for direct maximisation of the likelihood. The formulation can also be viewed as a sum of two terms: a reconstruction loss, and an aggregated posterior distribution matching loss term that is the KL divergence from the diagonal Gaussian prior. Wasserstein autoencoders (WAEs) proposed using divergences induced by the optimal transport problem, as an alternative to the KL divergence, for minimising the distance between the aggregated posterior and an independent Gaussian distribution. \cite{tolstikhinWassersteinAutoEncoders2018}. They achieved this in two different ways: by using a GAN to enforce distribution matching in the latent space, or by using a maximum mean discrepancy (MMD) loss term (with a choice of kernel function e.g. the inverse multi-quadratics (IMQ) kernel).

We propose a new adversarial jamming (AJ) approach to distribution matching based on a minimax game between a jammer, \(f(\cdot)\), and an auxiliary joint source channel coding (JSCC) encoder and decoder pair, \(g(\cdot)\) and \(h(\cdot)\), respectively, for a diagonal Gaussian source, \(X\), with a mean squared error (MSE) distortion measure for reconstruction under constraints on the power of both the jammer and the JSCC (see Figure~\ref{fig:jscc_scheme}). Existing theoretical results concerning the optimal (worst-case) additive jamming noise for such a setting \cite{akyolOptimalJammingAdditive2013, akyolOptimalJammingAdditive2013a, basarCompleteCharacterizationMinimax1985} imply that under these conditions, the optimal jammer output will be a diagonal Gaussian random vector to maximise the expected value of the reconstruction error. We exploit this fact in order to produce an adversarial jamming regularisation term for an autoencoder that tries to simultaneously minimise the distortion in compressing and reconstructing images from a dataset, while also matching a diagonal Gaussian prior. We show that this technique is effective for distribution matching in the latent space, comparable to a VAE and a WAE. To the best of our knowledge, this is a novel result that exploits a communication theoretic saddle point result to regularise the latent space of an autoencoder for distribution matching, which can lead to similar, potential future extensions.

\textbf{Notation.} We use \(\mathbf{0}\) and \(\mathbf{I}\) to represent the \(k\)-dimensional zero vector and \((k \times k)\)-dimensional identity matrix, respectively. Also, \(R_A = Q_A\Lambda_A Q_A^T\) denotes the eigenvalue decomposition of the covariance matrix \(R_A\) of random vector \(\mathbf{A}\) and \(F_A(\omega)\) denotes its characteristic function.
\section{Problem Definition}
 
Let \(\mathbf{X} \in \mathbb{R}^k\), \(\mathbf{D} \in \mathbb{R}^n\), and \(\mathbf{N} \in \mathbb{R}^k\) be random vectors representing the source distribution, jammer's source of randomness, and channel noise, respectively. Let \(f : \mathbb{R}^{n} \rightarrow \mathbb{R}^k\), \(g : \mathbb{R}^k \rightarrow \mathbb{R}^k\), and \(h : \mathbb{R}^k \rightarrow \mathbb{R}^k\) be Borel-measuarable functions. These respectively represent the jammer, with output \(\mathbf{Z} := f(\mathbf{D})\); the encoder (transmitter), with output \(\mathbf{Y} := g(\mathbf{X})\); and the decoder (receiver), with output \(\mathbf{\hat{X}} := h(\mathbf{\hat{Y}})\), where \(\mathbf{\hat{Y}}\) defines the additive noise channel mapping \(\mathbf{\hat{Y}} = \mathbf{Y} + \mathbf{N} + \mathbf{Z}\). Then, the zero-sum, minimax game among the transmitter, receiver, and adversarial jammer is shown below with cost function \(J(f, g, h)\), and power constraints \(P_t\) and \(P_a\) for the transmitter and jammer, respectively.

\begin{align}
    &\max_{f}\min_{h, g} J(f, g, h) \label{eq:minimax} \\
    &\text{where } J(f, g, h) = \mathbb{E}[(\mathbf{X} - \mathbf{\hat{X}})^2], \nonumber \\
    &\mathbb{E}[||g(\mathbf{X})||^2] \leq P_t \text{, and } \mathbb{E}[||f(\mathbf{D})||^2] \leq P_a \nonumber
\end{align}

The optimal combined policy for the transmitter, receiver, and jammer is the saddle point solution to the zero-sum game, \((f^*, h^*, g^*)\), where the two inequalities in Equation~\ref{eq:saddle_point_solution} are simultaneously satisfied.

\begin{align}
    J(f, h^*, g^*) \leq J(f^*, h^*, g^*) \leq J(f^*, h, g)
    \label{eq:saddle_point_solution}
\end{align}

It was shown in \cite{akyolOptimalJammingAdditive2013a} that the problem of optimal (zero-delay) jamming over a scalar additive noise channel (i.e. \(k=1\)) is closely connected to the linearity of \(g^*\) and \(h^*\). Where possible, a jammer will generate additive noise that forces the optimal transmitter and receiver to be linear transformations (the ``matching condition''). In the case of a scalar Gaussian source being sent over a Gaussian channel, the output of the jammer will also be Gaussian.

The scalar jamming result was extended and generalised in \cite{akyolOptimalJammingAdditive2013}, where it was shown that a vector version of the matching condition can be used to find saddle points for the vector version of the minimax game. An optimal encoding function for the transmitter will be a linear transformation (followed by a random sign change through multiplication by a Bernoulli random variable with \(p=0.5\) over the alphabet \(\{0, 1\}\)). It was shown that in the vector setting, the power allocation of the transmitter will obey reverse water-filling.

Our focus in this work is on the nature of the optimal jamming function. The necessary and sufficient condition for the linearity of optimal estimation - and a requirement for the optimal jamming strategy - is given in Equation~\ref{eq:matching_condition}, where \(\Sigma\) is a diagonal power allocation matrix. The optimal jammer also obeys water-filling, dependent on the eigenvalues of \(\mathbf{R_N}\) \cite{akyolOptimalJammingAdditive2013}.

\begin{align}
    \frac{\delta \log{F_{\Sigma Q_X^T \mathbf{X}} (\omega})}{\delta \omega_i} = S_i \frac{\delta \log{F_{Q_Z^T (\mathbf{N}+\mathbf{Z})} (\omega)}}{\delta \omega_i}, 1 \leq i \leq m
    \label{eq:matching_condition} \\
    \text{where } S = \Sigma \Lambda_X \Sigma \Lambda_Z^{-1}
    \nonumber
\end{align}

Such a characterisation of the nature of the optimal jamming solution implies that if \(\mathbf{X} \sim \mathcal{N}(\mathbf{0}, \mathbf{I})\) and \(\mathbf{N} \sim \mathcal{N}(\mathbf{0}, \mathbf{I})\), then the optimal jammer output will also be an isotropic diagonal Gaussian distribution \(\mathbf{Z} \sim \mathcal{N}(\mathbf{0}, \kappa \mathbf{I})\), where \(\kappa\) is a scalar constant, to match the source distribution and enforce the transmitter's linearity. This can be easily verified by noting that, in this special case, \(I = a\Sigma = b\Lambda_X = c\Lambda_Z = dQ_Z = eQ_X\), where \(a, b, c, d, e\) are scalar constants. Thus we have a solution \(\mathbf{Z} = \beta(\mathbf{N} - \mathbf{X})\), where \(\beta\) is scalar, which means that a saddle point solution for \(\mathbf{Z}\) is also a diagonal Gaussian random vector. This saddle point solution to the minimax game is also (almost surely) unique \cite{akyolOptimalJammingAdditive2013}.

\subsection{Jamming as Regularisation for an Autoencoder}

We propose the use of this communication theoretic game to regularise the latent space of an autoencoder. Consider a data distribution random vector, \(\mathbf{D} \in \mathbb{R}^n\); for example, flattened image data with \(n = l \cdot w \cdot m\), where \(l\), \(w\), and \(m\) are the length, width, and colour channels of the images, respectively. \(\mathbf{D}\) shall be the source of randomness for our jammer. We introduce an auxiliary DeepJSCC autoencoder that attempts to transmit and recover a source \(X \sim \mathcal{N}(\mathbf{0}, \mathbf{I})\) over the additive noise channel. DeepJSCC has been shown to be an effective technique for transmitting complex source distributions, such as natural images, over unknown communication channels \cite{bourtsoulatzeDeepJointSourceChannel2019, xuWirelessImageTransmission2021, yeChannelAgnosticEndtoEnd2018} and continues to be a topic that is actively researched in the field of semantic communication \cite{xuDeepJointSourceChannel2023}.

Since we deal with two autoencoders, one operating on data \(\mathbf{D}\) and one operating on Gaussian source  \(\mathbf{X}\), we use the terms ``data autoencoder'' and ``DeepJSCC autoencoder'' to distinguish between the two. We also refer to the encoder/decoder of each as the compressor/reconstructor and the transmitter/receiver, respectively. In the context of the vector jamming game as defined above, the reconstructor is \(f\), the transmitter is \(g\), and the receiver is \(h\). The reconstructor will be a function \(r : \mathbb{R}^k \rightarrow \mathbb{R}^n\) with output \(\mathbf{\hat{D}} := r(\mathbf{Z})\).

The objective of the DeepJSCC autoencoder is to minimise the reconstruction error of the Gaussian source, \(\mathcal{L}_{jscc}\), given in Equation~\ref{eq:jscc_objective}. In turn, the objective of the data autoencoder is to minimise an objective function, \(\mathcal{L}_{data}\), which is composed of two terms: its own reconstruction error for the data distribution, as well as the negative of \(\mathcal{L}_{jscc}\), as shown in Equation~\ref{eq:data_autoencoder_objective} below. Here, \(\mathcal{L}_{data}\) mirrors the objective of a VAE, with \(-\mathcal{L}_{jscc}\) replacing the usual KL divergence term.

\begin{align}
    \mathcal{L}_{jscc}(g, h) &= \mathbb{E}[(\mathbf{X}-\mathbf{\hat{X}})^2] \label{eq:jscc_objective} \\
    \mathcal{L}_{data}(f, r) &= \mathbb{E}[(\mathbf{D} - \mathbf{\hat{D}})^2] - \eta\mathbb{E}[(\mathbf{X}-\mathbf{\hat{X}})^2] 
    \label{eq:data_autoencoder_objective} 
\end{align}
where $\eta$ is a positive regularisation constant.

We enforce power constraints \(P_t = P_a = 1\) on the outputs \(\mathbf{Y}\) and \(\mathbf{Z}\) by normalising them (in practice, in mini-batch training with stochastic gradient descent, by using the empirical batch statistics). The competition between the compressor, acting as a jammer, and the DeepJSCC autoencoder, sets up a communication game in the form seen in Equation~\ref{eq:minimax}. The saddle point of this communication game is for the output of the jammer \(\mathbf{Z}\) to be an isotropic diagonal Gaussian. Our proposed approach makes use of the maximisation term from the game in Equation~\ref{eq:minimax} as a regularisation term that encourages the latent distribution to match this distribution. The proposed scheme is illustrated in Figure~\ref{fig:jscc_scheme}. For our experiments, we set the power of the fixed noise, \(\mathbf{N}\), to 0, using only the adversarial jammer output as noise.

\section{Experimental Results}

\begin{figure}[ht]
    \centering
    \includegraphics[width=\linewidth]{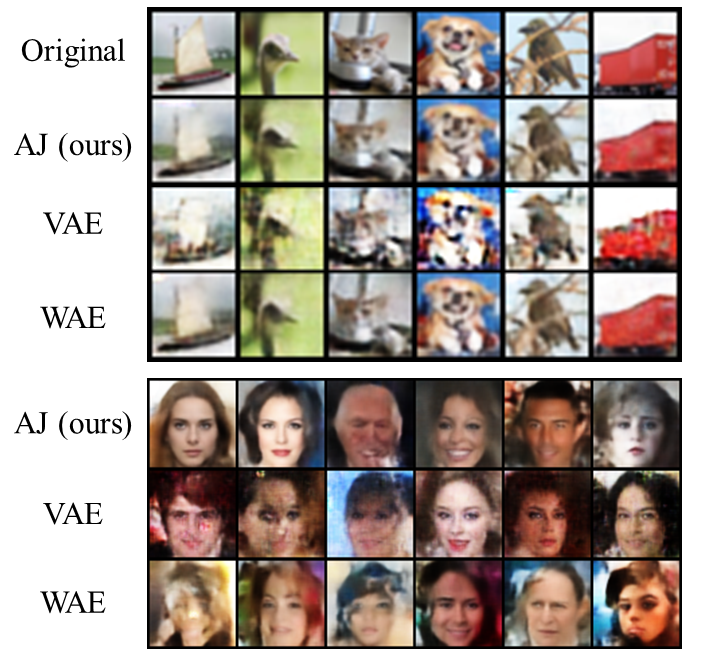}
    \caption{Reconstruction results for the CIFAR-10 dataset (top) and generation results for the CelebA dataset (bottom), for experiments using adversarial jamming, KL divergence (VAE), and maximum mean discrepancy (WAE) as the prior distribution latent regularisation term.}
    \label{fig:collated_figure}
\end{figure}

\begin{table*}[ht!]
\caption{FID scores, MSE losses for image reconstruction and the determinants of the Pearson correlation matrices (DPC). Arrows indicate whether higher (\(\uparrow\)) or lower (\(\downarrow\)) scores are better. These are shown for the three datasets. In each case, we show results for adversarial jamming, a standard VAE, and a WAE, trained using the same parameters and architecture. We see that the performance of adversarial jamming is comparable to WAEs and VAEs. \textbf{Boldface} entries denote the best performance.}
 \label{tab:scores}
\centering
\begin{tabular}{lllllllllllll}
\toprule
Dataset & \multicolumn{3}{c}{CelebA} & \multicolumn{3}{c}{CIFAR-10} & \multicolumn{3}{c}{MNIST} & \multicolumn{3}{c}{MNIST} \\
\midrule
Latent Dim & \multicolumn{3}{c}{64} & \multicolumn{3}{c}{64} & \multicolumn{3}{c}{2} & \multicolumn{3}{c}{8} \\
\midrule
Score & FID \(\downarrow\) & MSE  \(\downarrow\) & DPC  \(\uparrow\) & FID \(\downarrow\) & MSE  \(\downarrow\) & DPC  \(\uparrow\) & FID \(\downarrow\) & MSE  \(\downarrow\) & DPC  \(\uparrow\) & FID \(\downarrow\) & MSE  \(\downarrow\) & DPC  \(\uparrow\) \\
\midrule
AJ (ours) & 46.9 & 0.0065 & 0.50 & 91.1 & \textbf{0.0104} & 0.55 & 23.6 & \textbf{0.0131} & 0.96 & \textbf{85.8} & \textbf{0.0362} & \textbf{1.00} \\
VAE & 50.1 & 0.0065 & 0.02 & 105.3 & 0.0116 & 0.17 & 24.1 & 0.0133 & 0.78 & 94.7 & 0.0369 & 0.99\\
WAE & \textbf{44.2} & \textbf{0.0061} & \textbf{0.74} & \textbf{87.1} & 0.0105 & \textbf{0.75} & \textbf{22.0} & 0.0133 & \textbf{0.98} & 86.1 & 0.0370 & 1.00 \\
\bottomrule
\end{tabular}
\end{table*}

We use a DCGAN-like \cite{radfordUnsupervisedRepresentationLearning2016} convolutional autoencoder architecture for our data autoencoder, with output normalisation to ensure adherence to constraints on the variance and mean. We use a fully connected architecture that includes a linear mapping for our transmitter and receiver. The transmitter and receiver respectively have output and input shapes \(k\) (which is the transmitter output \(\mathbf{Y}\)), and \(k\) is also the length of the representation, which is the adversarial noise \(\mathbf{Z}\), output by the compressor. \(k\) is varied for each experiment, depending on the dataset. We test our method on three different datasets: CelebA \cite{liuDeepLearningFace2015}, CIFAR-10 \cite{krizhevskyLearningMultipleLayers2009}, and MNIST \cite{dengMNISTDatabaseHandwritten2012}. For CelebA we train for 75 epochs with \(k=64\), for CIFAR-10 we train for 250 epochs with \(k=64\), and for MNIST we run two experiments at \(k=2\) and \(k=8\), both for 250 epochs. We compare our method to a VAE trained using an ELBO loss term involving the KL divergence distribution matching, as well as a WAE with an MMD regularisation term using an IMQ Kernel. We use the same objective, parameters and architectures for these ablation experiments, changing only the distribution matching term.

\begin{figure}
    \centering
    \includegraphics[width=0.95\linewidth]{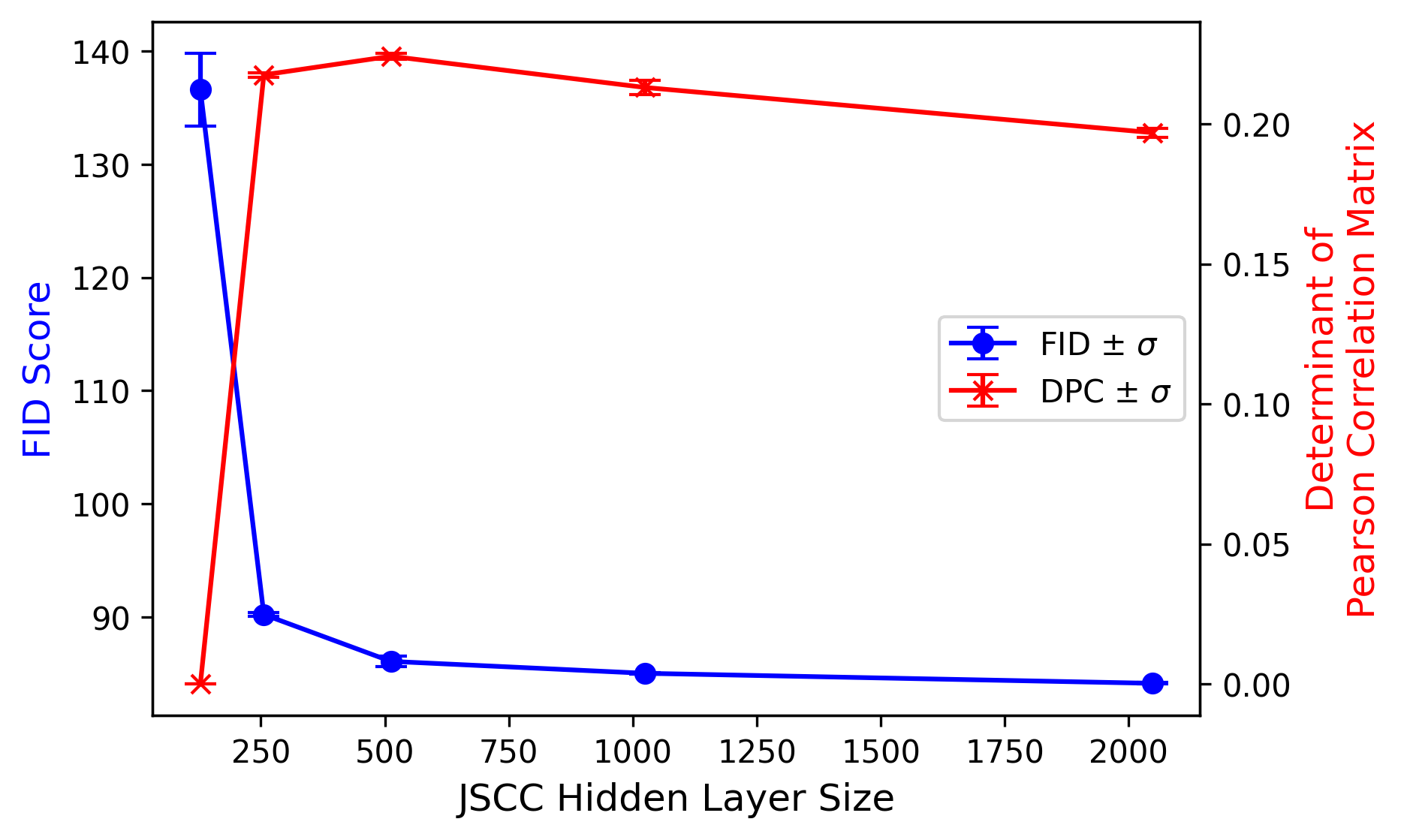}
    \caption{The FID score of a generative model trained using adversarial jamming versus the hidden layer size of the DeepJSSC autoencoder. Increasing the capacity of the DeepJSCC autoencoder for learning improves the FID score of the generative model.}
    \label{fig:fid_and_dpc_vs_hidden_size}
\end{figure}

Figure~~\ref{fig:collated_figure} shows the reconstruction results for CIFAR-10 and generation results for CelebA. Table~\ref{tab:scores} shows a comparison of three metrics for all the experiments. We measure the closeness of the distribution to an isotropic Gaussian indirectly, through two metrics: the Fréchet Inception Distance (FID) score \cite{heuselGANsTrainedTwo2017}, measuring the realism of the generated images; and the determinant of the Pearson correlation matrix (DPC) for the learned features, which is a measure of the degree of statistical independence between the learned features and thus the level of disentanglement of the features. Finally, the MSE is used to measure the fidelity of the image reconstruction. In Figure~\ref{fig:fid_and_dpc_vs_hidden_size}, we see a plot of the FID score and the determinant of the Pearson correlation matrix against the DeepJSCC autoencoder's hidden layer size for experiments conducted on the CIFAR-10 dataset for \(k=128\).

\section{Discussion}

Visually, the results in Figure~\ref{fig:collated_figure} demonstrate performance on par with a WAE in terms of reconstruction and generation. The high determinants of the Pearson correlation matrices in Table~\ref{tab:scores} suggest that our method may achieve a good level of feature independence, comparable to a VAE or WAE. This is reflected in the FID scores of the models, showing that when a diagonal Gaussian prior is used to sample new data, the output samples appear comparatively realistic. However, it should be noted that the choice of hyperparameters, e.g. the regularisation constant, may also affect the results and the trade-off between the MSE and FID.

In Figure~\ref{fig:fid_and_dpc_vs_hidden_size}, the hidden layer size is a way of parameterising the complexity/capacity of the autoencoder. We see that in general, the FID score decreases with increasing autoencoder capacity, and the opposite is approximately true for the determinant of the Pearson correlation matrices.
This can be interpreted intuitively as the fact that a more competitive DeepJSCC autoencoder is able to compete more effectively with the adversarial jammer and thus improves the prior distribution matching performance.

As an approach to impose a prior distribution on the latent space, adversarial jamming opens up the possibility for novel contexts in which competition can naturally promote the matching of the distribution, with no requirement for the explicit calculation of distribution divergences. A similar approach to distribution matching, where the Jensen-Shannon divergence is implicitly minimised through an adversarial game between a generator and a discriminator network, has already achieved success in the form of GANs \cite{goodfellowGenerativeAdversarialNetworks2014}.

It is also possible for a range of other distributions to be matched based on different non-Gaussian source distributions. We have seen that in the case of input random variables with isotropic covariance matrices, the jamming noise will be a linear combination of the source distribution and the channel noise distribution. It is therefore possible for the jammer to match the source distribution arbitrarily by specifying different source distributions for the DeepJSCC autoencoder. This will be explored as part of future work.

\section{Conclusion}

We have introduced adversarial jamming as a novel approach to distribution matching. This technique implicitly imposes distribution matching by encouraging an encoder to learn a mapping from a data distribution to adversarial jamming noise that disrupts the reconstruction task of an auxiliary joint source-channel coder. It increases the scope of possible settings in which distribution matching can be performed in the latent space of autoencoders and other neural networks. The technique has demonstrated good performance in the case of a diagonal Gaussian prior and could be extended to a range of other prior distributions, by considering different reconstruction problems involving alternative, non-Gaussian, source distributions.

\vfill\pagebreak

\bibliographystyle{IEEEbib}
\bibliography{references}

\end{document}